\documentclass[runningheads]{llncs}

\usepackage{eccv}

\usepackage{eccvabbrv}

\usepackage{graphicx}
\usepackage{booktabs}

\usepackage{multirow}
\usepackage{color}
\definecolor{best}{rgb}{1.0, 0.0, 0.0}
\definecolor{second}{rgb}{0.0, 0.0, 1.0}
\usepackage[accsupp]{axessibility}  

\usepackage{hyperref}

\usepackage{orcidlink}

\makeatletter
\newcommand{\corrauth}{\textsuperscript{\normalfont\@fnsymbol{1}}}
\makeatother

\begin{document}

\title{Learning to Balance: Decoupled Siamese Diffusion Transformer for Reference-Based Remote Sensing Image Super-Resolution}

\titlerunning{DS-DiT}

\author{
Bin Luo\inst{1} \and
Runmin Dong\inst{2}\thanks{Corresponding authors.} \and
Zhaoyang Luo\inst{1} \and
Jinxiao Zhang\inst{4} \and \\
Jiyao Zhao\inst{3} \and
Fan Wei\inst{4} \and
Haohuan Fu\inst{1,3,4}\corrauth
}
\authorrunning{B. Luo et al.}

\institute{Tsinghua Shenzhen International Graduate School, Shenzhen, China  \and
Sun Yat-sen University, Zhuhai, China
\and
National Supercomputing Center in Shenzhen, Shenzhen, China \and
Tsinghua University, Beijing, China
\\ 
\email{luob21@mails.tsinghua.edu.cn}, 
\email{dongrm3@mail.sysu.edu.cn}}

\maketitle

\begin{abstract}
Diffusion-based methods demonstrate significant potential for remote sensing image super-resolution at large scaling factors, particularly in reference-based super-resolution (RefSR), where high-resolution reference images provide critical fine-grained texture priors. However, existing methods often suffer from a trade-off between over-reliance on reference information, which leads to texture artifacts, and under-utilization of such information, which results in insufficient detail recovery. To address these issues, we propose DS-DiT, a Decoupled Siamese Diffusion Transformer that decouples the interaction between low-resolution (LR) and reference (Ref) conditions within the attention mechanism. By allowing LR structural priors and Ref texture information to independently interact with the noisy latent, the framework effectively mitigates competition between the two conditional sources. To further compensate for the limited local modeling ability of global attention, we introduce a Patch-Level Weighting (PLW) module that adaptively modulates the fusion of conditional sources. In addition, the siamese architecture enables an inference-time autoguidance strategy that exploits the prediction discrepancy between strong and weak Ref conditions to improve generation quality without additional training. Experimental results across multiple datasets and scaling factors show that DS-DiT outperforms existing methods in both quantitative metrics and visual fidelity. The source code is available at \url{https://github.com/B1nary-L/DS-DiT}.

\end{abstract}

\section{Introduction}
\label{sec:intro}

High-resolution (HR) remote sensing (RS) imagery is indispensable for a wide range of applications, including urban planning, environmental monitoring, and disaster assessment~\cite{dong2023large,jiang2020histif}. However, due to inherent constraints in sensor hardware and imaging conditions, acquired RS images often suffer from a trade-off between spatial and temporal resolutions~\cite{luo2018stair}, frequently failing to meet the requirements of fine-grained analysis. Image super-resolution (SR) techniques have emerged as an effective solution to bridge this gap by reconstructing HR images from easily accessible low-resolution (LR) observations with short revisit cycles, thereby effectively filling the temporal gaps in high-resolution data sequences.

Despite their utility, single-image super-resolution (SISR) methods face severe ill-posedness at large scaling factors~\cite{zhang2020texture}, such as $\times 8$ or $\times 16$, where they struggle to accurately recover lost high-frequency details. Reference-based super-resolution (RefSR) addresses this limitation by introducing an additional HR reference (Ref) image from the same geographic area but at a different acquisition time. The rich texture priors provided by these reference images effectively alleviate the information deficiency inherent in SISR. While LR and Ref images are typically multi-temporal observations of the same location and share aligned spatial extents, factors such as varying viewpoints, spectral differences, and land cover changes make the effective exploitation of reference features challenging. Consequently, early RefSR methods~\cite{zheng2018crossnet,jiang2021robust} focused primarily on feature alignment between LR and Ref to facilitate the transfer of reference textures.

The emergence of diffusion models~\cite{rombach2022high,zhang2023adding} has revolutionized RS RefSR by introducing multi-condition generative frameworks. Although diffusion models can generate remarkably rich details, especially by leveraging reference textures, they are prone to over-reliance on Ref features, which often leads to artifacts or false textures. Therefore, the core challenge in RS RefSR lies in effectively utilizing reference textures in unchanged regions while suppressing erroneous transfers in areas where land cover has changed. To this end, methods like Ref-Diff~\cite{dong2024building} introduce ideal land cover change priors as explicit conditions to guide the diffusion process. However, the performance of such methods degrades significantly in real-world applications where precise change detection masks are unavailable. Even the joint use of independent change detection and diffusion models often fails to produce stable and high-quality super-resolution results.

Meanwhile, recent advances in generative modeling have explored Multimodal Diffusion Transformer (MM-DiT) architectures~\cite{esser2024scaling,labs2025flux}, which enable effective cross-modal fusion through joint self-attention mechanisms. Nevertheless, for RS imagery, the standard text modality is insufficient for describing complex land cover features, rendering traditional text-guidance mechanisms ineffective. The optimal architectural design for integrating high-resolution reference images within the MM-DiT framework remains an open research challenge. A straightforward approach would be to replace the text branch in MM-DiT with LR and Ref images as new conditioning modalities. However, we observe that unconstrained interaction among these conditions leads to information competition, causing certain features to be over-exploited or under-utilized.

To mitigate these issues, and drawing inspiration from the architectural paradigms explored in CreatiLayout~\cite{zhang2025creatilayout} and TODSynth~\cite{yang2026task}, we propose DS-DiT, a Decoupled Siamese Diffusion Transformer that decouples the LR-Ref interaction within the attention mechanism. Unlike existing siamese designs that replicate noisy features, our design allows the noisy image tokens to generate a single shared set of query, key, and value (Q, K, and V) projections that are dispatched to two parallel joint attention paths. This configuration ensures that LR structural priors and Ref texture information interact independently yet consistently with the same noisy image latent, effectively alleviating inter-source competition while maintaining global coherence. Furthermore, since global attention mechanisms may overlook critical local details, we introduce a Patch-Level Weighting (PLW) module that adaptively regulates the fusion of multi-source information, enhancing fine-grained detail recovery.

Importantly, the siamese architecture design provides a consistent latent anchor for the parallel attention paths. This consistency allows us to introduce an autoguidance strategy during inference, which is analogous to classifier-free guidance~\cite{ho2022classifier}. By exploiting the prediction discrepancy of the same model under strong and weak reference conditions, we explicitly amplify reference utilization and improve generation quality without the need for additional training or auxiliary models. Extensive experiments on multiple remote sensing datasets across large scaling factors demonstrate that DS-DiT achieves superior results even in the absence of ideal change detection priors.

The main contributions of this work are summarized as follows:
\begin{itemize}
\item We propose a Decoupled Siamese Diffusion Transformer (DS-DiT), which mitigates inter-source information competition by decoupling the LR-Ref interaction within the attention mechanism. This decoupled architecture further enables an inference-time autoguidance strategy, which improves generation quality without requiring additional training or auxiliary models.
\item We design a Patch-Level Weighting (PLW) module to adaptively modulate the fusion of structural and textural features on a per-patch basis. By mitigating the detail loss inherent in global attention, this module enhances fine-grained detail recovery.
\item Extensive experiments on multiple remote sensing datasets demonstrate that DS-DiT outperforms state-of-the-art methods in both quantitative metrics and visual fidelity across diverse evaluation settings.
\end{itemize}

\section{Related Work}
\subsection{Reference-Based Image Super-Resolution}
Compared to single-image super-resolution (SISR), RefSR achieves higher-quality reconstruction by jointly exploiting a low-resolution (LR) image and an additional high-resolution reference (Ref) image. Early RefSR methods focus primarily on feature alignment between LR and Ref images to enable texture transfer. CrossNet~\cite{zheng2018crossnet} employs optical flow estimation for image alignment, while SRNTT~\cite{zhang2019image} performs multi-scale matching in the feature space. TTSR~\cite{yang2020learning} transfers textures via cross-attention, and AMSA~\cite{xia2022coarse} adopts a coarse-to-fine progressive matching strategy. C$^2$-Matching~\cite{jiang2021robust} leverages contrastive learning and knowledge distillation to handle cross-transformation and cross-resolution matching. DATSR~\cite{cao2022reference} further introduces deformable attention for more flexible feature alignment.

Unlike natural images, LR and Ref images in remote sensing scenarios are typically acquired from the same location at different times, and can thus be considered spatially aligned. Consequently, the core challenge of remote sensing RefSR lies in effectively exploiting reference textures in unchanged regions while suppressing the erroneous transfer of irrelevant textures in changed regions. However, existing remote sensing RefSR approaches~\cite{dong2021rrsgan, zhang2023reference} remain insufficient in utilizing effective texture information from Ref images. At large scaling factors, GAN-based methods struggle to faithfully reconstruct fine details in changed regions, leading to reconstructed images that lack high-frequency details. While recent diffusion-based works like Ref-Diff~\cite{dong2024building} attempt to use explicit land cover change priors to guide fusion, their reliance on additional change detection annotations limits their practical applicability. In contrast, our DS-DiT framework explores effective reference utilization without requiring any external change detection priors.

\subsection{Conditional Diffusion Models for Super-Resolution}

In recent years, diffusion models have achieved remarkable progress in image super-resolution owing to their powerful generative priors. SR3~\cite{saharia2022image} conditions on the LR image and performs super-resolution through iterative refinement. StableSR~\cite{wang2024exploiting} leverages the generative prior of a pretrained text-to-image diffusion model for blind super-resolution. DiffBIR~\cite{lin2024diffbir} adopts a two-stage strategy that first removes degradations and then supplements fine details via a diffusion model. PASD~\cite{yang2024pixel} introduces pixel-aware cross-attention, enabling the diffusion model to better perceive image structures. SeeSR~\cite{wu2024seesr} employs a degradation-aware prompt extractor to generate more realistic image details while preserving semantic consistency. DiT4SR~\cite{duan2025dit4sr} explores the capability of DiT-based diffusion models for real-world image super-resolution.

Despite their generative capability, diffusion models suffer from an inherent hallucination problem in SISR due to the lack of additional reference information, frequently producing textures inconsistent with the actual scene. Recent studies therefore combine RefSR with diffusion models, aiming to suppress hallucinations by leveraging Ref image information while generating high-quality images. CoSeR~\cite{sun2024coser} uses a diffusion model to synthesize a corresponding reference image for the LR input, and then injects both into the denoising process via ControlNet. ReFIR~\cite{guo2024refir} proposes a retrieval-augmented image restoration framework that exploits retrieved reference images to generate details faithful to the original scene. CRefDiff~\cite{wang2026controllable} fuses local and global information from the Ref image and leverages the generative prior of a diffusion model for real-world remote sensing image super-resolution. Ada-RefSR~\cite{wang2026trust} follows a trust-but-verify principle, adaptively exploiting reliable reference information while suppressing unreliable parts.

\subsection{Classifier-Free Guidance}
Classifier-Free Guidance (CFG)~\cite{ho2022classifier} is a widely used inference-time technique for improving generation quality by extrapolating between conditional and unconditional predictions. Applying CFG typically requires jointly training models by dropping conditions with a fixed probability. However, in our framework, the text branch has been removed, and simply replacing the Ref image with an unrelated image to simulate unconditional generation can degrade reference utilization during normal inference. To circumvent the need for separate unconditional training, techniques like ICG~\cite{sadat2025no} approximate unconditional predictions by sampling random independent conditions. Furthermore, Karras~\etal~\cite{karras2024guiding} suggest using a smaller or under-trained version of the model in place of the unconditional model for guidance. Inspired by these self-guidance strategies, we propose an autoguidance strategy specifically tailored for the decoupled siamese architecture. This strategy leverages the prediction discrepancy between strong and weak reference conditions to explicitly amplify reference utilization without additional training.

\begin{figure}[tb]
  \centering
  \includegraphics[width=\textwidth]{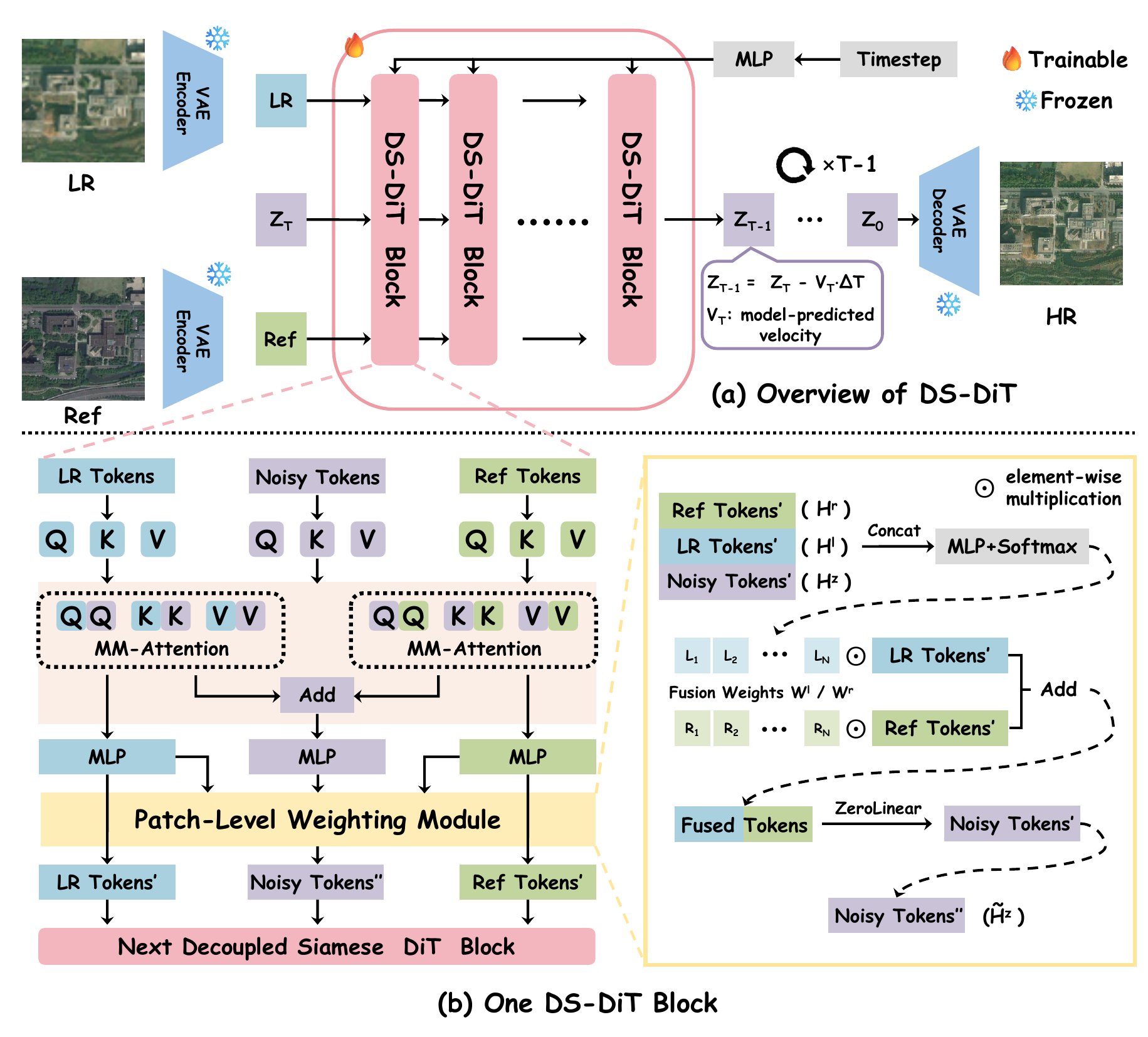}
  \caption{Overall architecture of the proposed Decoupled Siamese Diffusion Transformer (DS-DiT). Our framework employs a decoupled siamese interaction with a shared set of Q, K, and V to mitigate inter-source competition, complemented by a Patch-Level Weighting (PLW) module for adaptive local feature fusion.}
  \label{fig:framework}
\end{figure}

\section{Methodology}

In this work, we adopt conditional flow matching as the training paradigm to improve the reconstruction quality of RefSR at large scaling factors. The overall workflow of the proposed method is shown in \cref{fig:framework}. We modify the MM-DiT architecture and design a Decoupled Siamese Diffusion Transformer, which introduces LR and Ref images as conditioning modalities on equal footing. In the joint attention mechanism, the guidance from LR and Ref branches is explicitly decoupled to prevent direct interaction between them, thereby alleviating inter-source competition. Furthermore, to capture local information that is easily overlooked by global attention, we adaptively fuse LR and Ref information at the patch level and inject the fused features into the denoising branch. During inference, we exploit the prediction discrepancy of the same model under different reference conditioning strengths to guide the denoising direction, further improving reconstruction quality without any additional training.

\subsection{Preliminary}
\label{subsec:preliminary}

Unlike early diffusion models~\cite{ho2020denoising,song2020denoising} that rely on stochastic differential equations (SDEs) to describe the diffusion process, Flow Matching~\cite{lipman2022flow} directly models sample trajectories using ordinary differential equations (ODEs), resulting in more efficient training and faster sampling. Rectified Flow~\cite{liu2022flow} further simplifies this modeling process by constraining the transport paths between samples to linear interpolation trajectories:
\begin{equation}
  \mathbf{x}_t = (1-t)\, \mathbf{x}_0 + t\, \mathbf{x}_1, \quad t \in [0,1].
  \label{eq:rectified_flow}
\end{equation}
The training objective is to let the model-predicted velocity field $\mathbf{v}_t^\theta$ fit the ground-truth velocity field $\mathbf{v} = \mathrm{d}\mathbf{x}_t / \mathrm{d}t = \mathbf{x}_1 - \mathbf{x}_0$:
\begin{equation}
  \mathcal{L}_{\mathrm{RF}}(\theta) = \mathbb{E}_{t, \mathbf{x}_0, \mathbf{x}_1} \left\| \mathbf{v}_t^\theta - (\mathbf{x}_1 - \mathbf{x}_0) \right\|_2^2.
  \label{eq:rf_loss}
\end{equation}

Current mainstream text-to-image models~\cite{esser2024scaling,stability2024sd35,blackforest2024flux} adopt the MM-DiT architecture to predict the velocity field. Previous diffusion models typically rely on cross-attention to fuse image and text information, whereas MM-DiT designs independent Transformer branches for each modality. After encoding image and text features in their respective feature spaces, MM-DiT concatenates the resulting tokens along the sequence dimension and performs joint self-attention. This design enables image and text features to interact fully within a unified representation space, yielding superior multimodal fusion. Inspired by this, we treat LR and Ref images as conditioning modalities and model them in a manner similar to the text modality in MM-DiT, enabling the model to exploit structural priors from the LR image and texture information from the Ref image for high-quality reconstruction.

\subsection{Decoupling LR and Ref Interaction in MM-Attention}
\label{subsec:siamese_attention}

The text modality is unable to accurately describe land cover features in low-resolution remote sensing images, making it ineffective for remote sensing super-resolution tasks. We therefore completely discard the text branch and model LR and Ref as new conditioning modalities. A straightforward idea is to construct an M$^3$-DiT architecture that allows unconstrained interactions among all three modalities in the attention mechanism. Details of M$^3$-DiT are provided in \cref{sec:appendix_m3dit} of the appendix. However, in this setting, the softmax operation in self-attention forces information from different sources to compete for limited attention weights within a single unified sequence, leading to feature dilution where structural priors or texture information are under-utilized. To alleviate this issue, we adopt a decoupled siamese architecture that separates the interaction between LR and Ref conditions.

In CreatiLayout~\cite{zhang2025creatilayout}, the image branch generates independent sets of queries (Q), keys (K), and values (V) for the joint attention with each modality. In contrast, in our design, the noisy image tokens $\mathbf{h}^z$ produce only a single set of Q, K, and V in each block, which is shared by both LR and Ref branches. The interaction between the LR tokens and the noisy image tokens is realized through MM-Attention:
\begin{equation}
    \mathbf{h}_l^z,\; \mathbf{h}_{\mathrm{new}}^l = \mathrm{Attention}\!\left(
        [\mathbf{h}^z \mathbf{W}_q^z,\, \mathbf{h}^l \mathbf{W}_q^l],\;
        [\mathbf{h}^z \mathbf{W}_k^z,\, \mathbf{h}^l \mathbf{W}_k^l],\;
        [\mathbf{h}^z \mathbf{W}_v^z,\, \mathbf{h}^l \mathbf{W}_v^l]
    \right),
    \label{eq:lr_attention}
\end{equation}
where $[\cdot,\cdot]$ denotes concatenation along the sequence dimension, $\mathbf{W}_q^z, \mathbf{W}_k^z, \mathbf{W}_v^z$ are the learnable projection matrices for the noisy image branch, $\mathbf{W}_q^l, \mathbf{W}_k^l, \mathbf{W}_v^l$ are the corresponding projection matrices for the LR branch, and $\mathbf{h}_l^z$ and $\mathbf{h}_{\mathrm{new}}^l$ are the updated noisy image tokens and LR tokens after interaction, respectively. Meanwhile, the Ref tokens and the noisy image tokens undergo the same joint attention mechanism:
\begin{equation}
    \mathbf{h}_r^z,\; \mathbf{h}_{\mathrm{new}}^r = \mathrm{Attention}\!\left(
        [\mathbf{h}^z \mathbf{W}_q^z,\, \mathbf{h}^r \mathbf{W}_q^r],\;
        [\mathbf{h}^z \mathbf{W}_k^z,\, \mathbf{h}^r \mathbf{W}_k^r],\;
        [\mathbf{h}^z \mathbf{W}_v^z,\, \mathbf{h}^r \mathbf{W}_v^r]
    \right).
    \label{eq:ref_attention}
\end{equation}
The updated noisy image tokens are then obtained via a summation operation:
\begin{equation}
    \mathbf{h}_{\mathrm{new}}^z = \mathbf{h}_l^z + \mathbf{h}_r^z.
    \label{eq:token_sum}
\end{equation}
In this way, the guidance from LR and Ref is explicitly decoupled at the attention stage: each interacts with the noisy image independently, effectively mitigating inter-source competition.

\subsection{Injecting LR and Ref Information via Patch-Level Weighting}
\label{subsec:patch_weights}

In the DiT architecture, the attention mechanism operates at the global level. However, relying solely on global attention may lead to the loss of local information~\cite{duan2025dit4sr}, which is equally critical for recovering fine-grained image details. It is therefore necessary to strengthen the fusion of LR and Ref features at the local level. Notably, the information provided by Ref is not entirely reliable, as land cover in certain regions may have changed between the acquisition times of LR and Ref. For such changed regions, the guidance from LR should be strengthened to maintain fidelity, whereas for regions where land cover remains unchanged, enhancing the guidance from Ref is more beneficial for recovering fine textures. Based on this analysis, we adaptively fuse LR and Ref through Patch-Level Weighting and inject the fused features into the denoising branch.

Specifically, after the joint attention computation, the LR, Ref, and noisy image tokens $\mathbf{h}_{\mathrm{new}}^l$, $\mathbf{h}_{\mathrm{new}}^r$, and $\mathbf{h}_{\mathrm{new}}^z$ are each passed through their respective MLPs, yielding feature tokens $\mathbf{H}^l, \mathbf{H}^r, \mathbf{H}^z \in \mathbb{R}^{N \times C}$. We first concatenate these three along the feature dimension and feed the result into a three-layer MLP, which outputs a per-patch weight map $\mathbf{W} \in \mathbb{R}^{N \times 2}$. A softmax operation is applied along the last dimension of $\mathbf{W}$ to assign specific fusion weights for each patch, and the resulting weight map is then split into LR and Ref weights. The weighted LR and Ref features are then fused and injected into the noisy tokens as follows:
\begin{equation}
    \tilde{\mathbf{H}}^z = \mathbf{H}^z + \mathrm{ZeroLinear}\!\left( \mathbf{W}^l \odot \mathbf{H}^l + \mathbf{W}^r \odot \mathbf{H}^r \right),
    \label{eq:patch_fusion}
\end{equation}
where $\mathbf{W}^l$ and $\mathbf{W}^r$ are the per-patch weights for LR and Ref, respectively, $\mathrm{ZeroLinear}$ denotes a linear projection layer with zero-initialized weights to ensure training stability, $\odot$ denotes element-wise multiplication with broadcasting, and $\tilde{\mathbf{H}}^z$ is the final output of the denoising branch in the current block.

\subsection{Improving Image Quality with Autoguidance}
\label{subsec:autoguidance}

In conditional rectified flow, the ideal velocity field $\mathbf{v} = \mathbf{x}_1 - \mathbf{x}_0$ is a constant that depends only on the initial noise $\mathbf{x}_1$ and the target data $\mathbf{x}_0$. In the RefSR task, however, given LR and Ref as two conditions, the model-predicted velocity field $\mathbf{v}_t^\theta(\mathbf{x}_t \mid c_{\mathrm{lr}}, c_{\mathrm{ref}})$ deviates from the ideal velocity field and varies with the current state $\mathbf{x}_t$. If the predicted velocity field can be corrected through guidance at each sampling step, the entire sampling trajectory can be optimized toward the target distribution, leading to higher-quality reconstruction.

In our siamese attention mechanism, the noisy image tokens are updated via $\mathbf{h}_{\mathrm{new}}^z = \mathbf{h}_l^z + \mathbf{h}_r^z$. During sampling, we introduce a scaling coefficient $\lambda$ for the tokens produced by the reference interaction path:
\begin{equation}
    \mathbf{h}_{\mathrm{new}}^z = \mathbf{h}_l^z + \lambda\, \mathbf{h}_r^z.
    \label{eq:lambda_scaling}
\end{equation}
When $\lambda = 1$, the model predicts the velocity field $\mathbf{v}_t^\theta(\mathbf{x}_t \mid c_{\mathrm{lr}}, c_{\mathrm{ref}})$ in its original direction. To construct a weak reference condition, we set $\lambda = 0$, which eliminates the direct contribution of the Ref branch to the noisy tokens and significantly reduces the guidance strength. The model then predicts a weakened velocity field $\mathbf{v}_t^{\theta^-}(\mathbf{x}_t \mid c_{\mathrm{lr}}, c_{\mathrm{ref}}^-)$. Analogous to the extrapolation form of CFG, the corrected velocity field $\mathbf{v}_t^{\theta^+}$ can be expressed as:
\begin{equation}
    \mathbf{v}_t^{\theta^+}(\mathbf{x}_t \mid c_{\mathrm{lr}}, c_{\mathrm{ref}}^+) = (1 - \omega)\, \mathbf{v}_t^{\theta^-}(\mathbf{x}_t \mid c_{\mathrm{lr}}, c_{\mathrm{ref}}^-) + \omega\, \mathbf{v}_t^\theta(\mathbf{x}_t \mid c_{\mathrm{lr}}, c_{\mathrm{ref}}),
    \label{eq:autoguidance}
\end{equation}
where $\omega$ is the guidance coefficient. When $\omega > 1$, the prediction trajectory is extrapolated along the reference guidance direction, shifting toward more thorough exploitation of the texture information from Ref. It should be noted that, to preserve the structural fidelity of the reconstructed image, $\omega$ must be kept within a moderate range to prevent the prediction from departing from the structural constraints imposed by the LR image. A detailed analysis of the influence of $\omega$ is provided in \cref{sec:appendix_autoguidance} of the appendix.

\section{Experiments}
\label{sec:experiments}

\subsection{Experimental Settings}
\label{subsec:settings}

\noindent\textbf{Datasets.}
SECOND~\cite{yang2020semantic} is a semantic change detection dataset comprising 4,662 pairs of aerial images. Each image has a size of $512 \times 512$ pixels with a spatial resolution ranging from 0.5 to 3 meters. The dataset covers six major land cover categories across diverse sensors and regions, encompassing various imaging styles and scene types. We follow the official split~\cite{yang2020semantic}, using 2,968 image pairs for training and 1,694 pairs for testing.

FUSU~\cite{yuan2024fusu} is a high-resolution aerial land cover change detection dataset sourced from Google Earth and annotated with 17 land cover categories. The cropped images are $512 \times 512$ in size with a spatial resolution of 0.2 to 0.5 meters. We select 7,436 image pairs for training and 2,192 pairs for testing.

CNAM-CD~\cite{zhou2023signet} is a remote sensing change detection dataset collected from Google Earth, with a spatial resolution of 0.5~m. In our experiments, we use 1,758 image pairs for training and 750 image pairs for testing. 

Real-RefRSSRD~\cite{wang2026controllable} is a real-world remote sensing RefSR dataset. In contrast to synthetic-degradation settings, Real-RefRSSRD contains paired HR and LR satellite images collected from different sensors. Specifically, the HR images are from NAIP with a ground sampling distance (GSD) of 1~m, while the LR images are from Sentinel-2 with a GSD of 10~m. This cross-sensor setting provides a more realistic evaluation scenario.

\noindent\textbf{Implementation Details.}
For SECOND, FUSU, and CNAM-CD, we conduct super-resolution experiments at two challenging scaling factors: $\times 8$ and $\times 16$. Low-resolution (LR) images are generated via bicubic downsampling. While we do not simulate complex real-world degradations, these large scaling factors themselves pose significant challenges for structural and textural recovery. For Real-RefRSSRD, we follow its original real-world setting and perform $\times 10$ super-resolution. DS-DiT is built upon the pretrained MM-DiT blocks of SD3~\cite{esser2024scaling}. Specifically, we extract the denoising branch weights to initialize the dual siamese branches for stable training. For each dataset and scale setting, the model is trained at the target resolution, i.e., $512 \times 512$ for SECOND, FUSU, and CNAM-CD, and $480 \times 480$ for Real-RefRSSRD. LR images are upsampled to the corresponding target resolution using bicubic interpolation before being fed into the model. We utilize the AdamW~\cite{loshchilov2017decoupled} optimizer with a learning rate of $5 \times 10^{-5}$ and a weight decay of 0.001. Each model is trained on 4 NVIDIA A100 GPUs with a total batch size of 16 for 80k steps, taking approximately 21 hours. For inference, we use the Euler ODE solver with 40 steps. The guidance coefficient $\omega$ is set to 1.2 for SECOND and 1.1 for the remaining datasets.

\begin{table}[t]
  \caption{Quantitative comparison of RefSR methods on the SECOND and FUSU datasets at $\times 8$ and $\times 16$ scaling factors. {\color{best}Red} indicates the best and {\color{second}blue} indicates the second-best result. (Creati$^{*}$ denotes CreatiLayout$^{*}$.)
  }
  \label{tab:quantitative}
  \centering
  \begin{tabular}{@{}ccc ccccccc@{}}
    \toprule
    Dataset & Scale & Metric & TTSR & DATSR & CoSeR$^*$ & Ref-Diff & Creati$^*$ & M$^3$-DiT & Ours \\
    \midrule
    \multirow{10}{*}{SECOND} & \multirow{5}{*}{$\times 8$} 
      & LPIPS~$\downarrow$  & 0.2994 & 0.3835 & 0.2576 & 0.2819 & 0.2692 & {\color{second}0.2417} & {\color{best}0.2174} \\
    & & FID~$\downarrow$    & 34.62  & 40.87  & 23.06  & 28.48  & 24.05  & {\color{second}22.66}  & {\color{best}15.87} \\
    & & DISTS~$\downarrow$  & 0.1518 & 0.1923 & {\color{second}0.1289} & 0.1490 & 0.1412 & 0.1356 & {\color{best}0.1090} \\
    & & CLIPIQA~$\uparrow$ & 0.2937 & 0.2076 & 0.2973 & 0.4361 & 0.4255 & {\color{second}0.4434} & {\color{best}0.4698} \\
    & & MUSIQ~$\uparrow$   & 43.86  & 27.36  & 42.11  & {\color{best}51.36}  & 43.18  & 44.92  & {\color{second}49.04} \\
    \cmidrule{2-10}
    & \multirow{5}{*}{$\times 16$} 
      & LPIPS~$\downarrow$  & 0.3927 & 0.5039 & {\color{second}0.3426} & 0.3503 & 0.3556 & 0.3504 & {\color{best}0.2905} \\
    & & FID~$\downarrow$    & 134.68 & 99.74  & 31.12  & {\color{second}30.11}  & 32.03  & 35.50  & {\color{best}21.58} \\
    & & DISTS~$\downarrow$  & 0.2196 & 0.2555 & 0.1595 & {\color{second}0.1555} & 0.1670 & 0.1715 & {\color{best}0.1334} \\
    & & CLIPIQA~$\uparrow$ & 0.3391 & 0.1554 & 0.3134 & 0.3468 & {\color{second}0.4530} & 0.4489 & {\color{best}0.4875} \\
    & & MUSIQ~$\uparrow$   & 37.71  & 22.02  & 42.18  & {\color{best}49.75}  & 43.55  & 43.65  & {\color{second}48.65} \\
    \midrule
    \multirow{10}{*}{FUSU} & \multirow{5}{*}{$\times 8$} 
      & LPIPS~$\downarrow$  & 0.2960 & 0.3295 & 0.2447 & 0.2809 & {\color{second}0.2143} & 0.2167 & {\color{best}0.1741} \\
    & & FID~$\downarrow$    & 54.73  & 35.06  & 24.69  & 26.21  & {\color{second}16.88}  & 17.95  & {\color{best}13.01} \\
    & & DISTS~$\downarrow$  & 0.1660 & 0.1803 & 0.1346 & 0.1618 & {\color{second}0.1282} & 0.1289 & {\color{best}0.1054} \\
    & & CLIPIQA~$\uparrow$ & 0.4998 & 0.2623 & 0.4178 & {\color{best}0.5495} & 0.5076 & 0.5095 & {\color{second}0.5306} \\
    & & MUSIQ~$\uparrow$   & 51.33  & 32.82  & 49.54  & {\color{best}54.43}  & 51.45  & 51.45  & {\color{second}52.84} \\
    \cmidrule{2-10}
    & \multirow{5}{*}{$\times 16$} 
      & LPIPS~$\downarrow$  & 0.3398 & 0.4546 & 0.2954 & 0.3288 & 0.2766 & {\color{second}0.2641} & {\color{best}0.2301} \\
    & & FID~$\downarrow$    & 127.12 & 104.78 & 35.01  & 34.26  & 20.55  & {\color{second}20.35}  & {\color{best}19.52} \\
    & & DISTS~$\downarrow$  & 0.1949 & 0.2493 & 0.1652 & 0.1714 & 0.1510 & {\color{second}0.1456} & {\color{best}0.1309} \\
    & & CLIPIQA~$\uparrow$ & 0.4836 & 0.2360 & 0.4036 & {\color{second}0.5175} & 0.5066 & 0.5101 & {\color{best}0.5339} \\
    & & MUSIQ~$\uparrow$   & 48.85  & 27.72  & 49.11  & {\color{second}53.00}  & 50.78  & 51.55  & {\color{best}53.21} \\
    \bottomrule
  \end{tabular}%
\end{table}

\begin{table}[tb]
  \caption{Quantitative comparison on CNAM-CD and Real-RefRSSRD. {\color{best}Red} indicates the best and {\color{second}blue} indicates the second-best result. (Creati$^{*}$ denotes CreatiLayout$^{*}$.)}
  \label{tab:other_datasets}
  \centering
  \begin{tabular}{@{}ccc ccccc@{}}
    \toprule
    Dataset & Scale & Metric & CoSeR$^{*}$ & Ref-Diff & Creati$^{*}$ & M$^3$-DiT & Ours \\
    \midrule
    \multirow{10}{*}{CNAM-CD}
    & \multirow{5}{*}{$\times 8$}
     & LPIPS~$\downarrow$   & 0.2593 & 0.2787 & \textcolor{second}{0.2560} & 0.2566 & \textcolor{best}{0.2101} \\
    & & FID~$\downarrow$     & 51.00  & 51.97  & \textcolor{second}{45.92} & 46.50 & \textcolor{best}{35.76} \\
    & & DISTS~$\downarrow$   & \textcolor{second}{0.1365} & 0.1465 & 0.1434 & 0.1441 & \textcolor{best}{0.1109} \\
    & & CLIPIQA~$\uparrow$   & 0.3322 & 0.3794 & \textcolor{second}{0.4058} & 0.3980 & \textcolor{best}{0.4446} \\
    & & MUSIQ~$\uparrow$     & 36.85  & \textcolor{best}{47.15} & 39.71 & 40.06 & \textcolor{second}{44.08} \\
    \cmidrule(lr){2-8}
    & \multirow{5}{*}{$\times 16$}
     & LPIPS~$\downarrow$   & \textcolor{second}{0.3319} & 0.3405 & 0.3506 & 0.3486 & \textcolor{best}{0.2789} \\
    & & FID~$\downarrow$     & 62.99 & \textcolor{second}{60.76} & 72.05 & 64.84 & \textcolor{best}{45.79} \\
    & & DISTS~$\downarrow$   & 0.1716 & \textcolor{second}{0.1545} & 0.1915 & 0.1827 & \textcolor{best}{0.1361} \\
    & & CLIPIQA~$\uparrow$   & 0.3301 & 0.3284 & 0.4019 & \textcolor{second}{0.4052} & \textcolor{best}{0.4325} \\
    & & MUSIQ~$\uparrow$     & 34.79 & \textcolor{best}{46.25} & 36.82 & 38.97 & \textcolor{second}{43.35} \\
    \midrule
    \multirow{5}{*}{Real-RefRSSRD}
    & \multirow{5}{*}{$\times 10$}
     & LPIPS~$\downarrow$   & 0.4459 & 0.3979 & 0.3825 & \textcolor{second}{0.3756} & \textcolor{best}{0.3261} \\
    & & FID~$\downarrow$     & 52.33  & 48.20  & \textcolor{second}{34.64} & 35.18 & \textcolor{best}{31.56} \\
    & & DISTS~$\downarrow$   & 0.2450 & \textcolor{second}{0.2067} & 0.2192 & 0.2072 & \textcolor{best}{0.1809} \\
    & & CLIPIQA~$\uparrow$   & 0.4818 & 0.5870 & 0.6281 & \textcolor{second}{0.6299} & \textcolor{best}{0.6456} \\
    & & MUSIQ~$\uparrow$     & 46.07  & 56.43  & \textcolor{second}{57.07} & 56.90 & \textcolor{best}{57.09} \\
    \bottomrule
  \end{tabular}
\end{table}

\noindent\textbf{Evaluation Metrics.}
To quantitatively evaluate RefSR performance, we adopt three reference-based perceptual metrics. LPIPS~\cite{zhang2018perceptual} measures perceptual similarity between images using features extracted by a pretrained deep network. FID~\cite{heusel2017gans} computes the distributional distance between generated and real images in feature space. DISTS~\cite{ding2020iqa} evaluates structural and textural similarity. Compared to PSNR and SSIM, these metrics are more suitable for assessing generative super-resolution methods~\cite{yang2024pixel}. In addition, we introduce two no-reference metrics, CLIPIQA~\cite{wang2023exploring} and MUSIQ~\cite{ke2021musiq}, to comprehensively evaluate the visual quality of reconstructed images.

\subsection{Comparison Results}
\label{subsec:comparison}

For SECOND and FUSU, we compare the proposed method with state-of-the-art GAN-based and diffusion-based RefSR methods. These include two GAN-based methods (TTSR~\cite{yang2020learning} and DATSR~\cite{cao2022reference}), two U-Net-based diffusion methods (CoSeR$^*$~\cite{sun2024coser} and Ref-Diff~\cite{dong2024building}), and two DiT-based methods (CreatiLayout$^*$~\cite{zhang2025creatilayout} and M$^3$-DiT introduced in \cref{subsec:siamese_attention}). For CNAM-CD and Real-RefRSSRD, we compare our method with the diffusion-based methods. Specifically, CoSeR$^*$ denotes an adapted version of the original method, in which the automatically generated reference image is replaced with the paired real reference image from the dataset. CreatiLayout$^*$ is likewise adapted for the RefSR task by replacing the original text and layout inputs with LR and Ref, and adopting a two-stage training strategy: a SISR model is first trained, after which most parameters are frozen and the Ref branch is introduced for continued training.

\begin{figure}[tb]
  \centering
  \includegraphics[width=\textwidth]{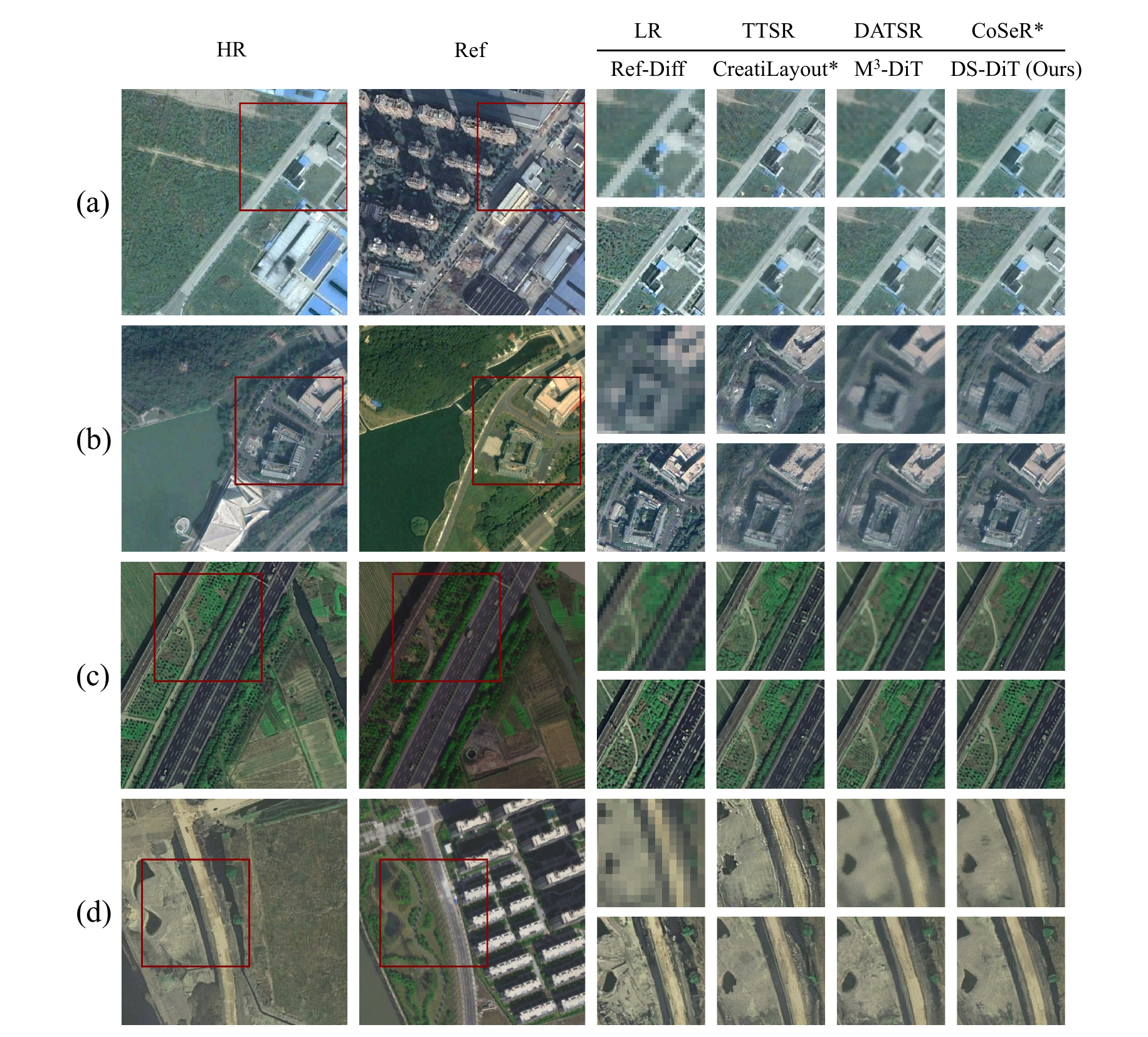}
  \caption{Qualitative comparison of different RefSR methods on the SECOND and FUSU datasets. 
    The visual results correspond to: (\emph{a}) SECOND at $\times 8$ scaling factor, (\emph{b}) SECOND at $\times 16$ scaling factor, (\emph{c}) FUSU at $\times 8$ scaling factor, and (\emph{d}) FUSU at $\times 16$ scaling factor. Please zoom in for better visual comparison.
  }
  \label{fig:qualitative}
\end{figure}

\noindent\textbf{Quantitative Comparison.}
\cref{tab:quantitative} presents the results on SECOND and FUSU at two large scaling factors ($\times 8$ and $\times 16$). The proposed method achieves the best performance on all reference-based metrics, demonstrating that our results possess both high fidelity and perceptual quality. On the no-reference metrics, our method also shows strong competitiveness, falling slightly behind Ref-Diff only in certain settings. As shown in \cref{fig:qualitative}, Ref-Diff tends to generate visually rich yet less faithful textures in some regions. Since no-reference metrics such as CLIPIQA~\cite{wang2023exploring} and MUSIQ~\cite{ke2021musiq} primarily reflect perceptual quality rather than explicit fidelity to the ground truth, such rich but non-authentic details may still receive higher scores. In contrast, our method achieves a better balance between reconstruction fidelity and perceptual realism.

\cref{tab:other_datasets} reports the results on CNAM-CD and Real-RefRSSRD. On CNAM-CD, our method consistently achieves the best results on all reference-based metrics at both $\times 8$ and $\times 16$ scaling factors. On Real-RefRSSRD, our method achieves the best performance across all metrics under the real-world $\times 10$ setting. These results show that the proposed method remains effective across different data distributions and realistic cross-sensor degradation scenarios.

\noindent\textbf{Qualitative Comparison.}
\cref{fig:qualitative} provides visual comparison results on SECOND and FUSU. \cref{fig:qualitative}(a) shows an example from the SECOND $\times 8$ dataset with substantial land cover changes. Even though only a few regions remain unchanged, our method successfully extracts the corresponding textures and reconstructs the fine details of the buildings on the right side of the image. \cref{fig:qualitative}(b) presents an example from the SECOND $\times 16$ dataset with minor changes. At this extreme scaling factor, other methods fail to transfer textures successfully, producing artifacts or blurry results despite the valuable information provided by Ref. In contrast, our method captures usable texture information and transfers it effectively, recovering the complex structures of the buildings. \cref{fig:qualitative}(c) illustrates an example from the FUSU $\times 8$ dataset. When the vegetation information in Ref is untrustworthy due to seasonal changes, our method avoids blindly transferring erroneous textures; meanwhile, it effectively exploits reliable road textures to assist reconstruction---a selective transfer capability that other methods lack. Finally, \cref{fig:qualitative}(d) shows an example from the FUSU $\times 16$ dataset with drastic changes, where our method still successfully recovers the clear shapes of the lake and the river. In summary, the qualitative comparisons further validate the effectiveness of the proposed method, which selectively exploits Ref information rather than blindly copying it. Under the structural constraints imposed by the LR image, our approach generates reconstructed results with higher fidelity and superior perceptual quality. More visual comparison results are provided in the appendix.

\subsection{Ablation Studies}
\label{subsec:ablation}

We conduct ablation studies on the SECOND dataset at both $\times 8$ and $\times 16$ scaling factors to validate the effectiveness of each component. Detailed results are presented in \cref{tab:ablation}.

\begin{figure}[tb]
  \centering
  \includegraphics[width=\textwidth]{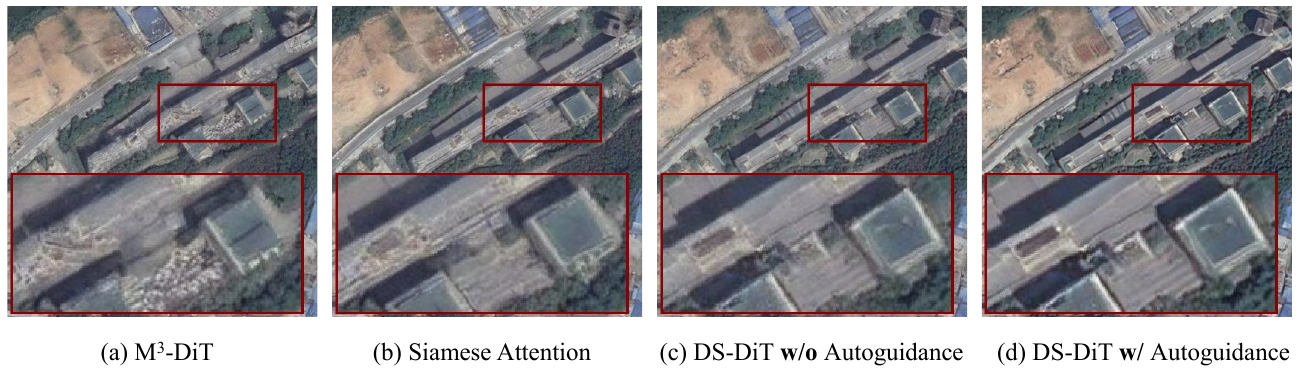}
    \caption{Visual ablation study on the SECOND dataset ($\times 16$ scaling factor).}
  \label{fig:ablation}
\end{figure}

\noindent\textbf{Importance of Decoupled Siamese Attention.} 
Compared to M$^3$-Attention, our decoupled siamese attention separates the interaction between LR and Ref conditions. Although this avoids their explicit interaction within the same attention layer, it achieves superior performance on nearly all metrics (see the comparison between M$^3$-DiT in \cref{tab:quantitative} and the siamese attention baseline in \cref{tab:ablation}). This confirms that by mitigating inter-source competition, the decoupled architecture improves information utilization and prevents feature dilution. As observed in \cref{fig:ablation}(a) and \cref{fig:ablation}(b), this architecture recovers more building and ground details compared to M$^3$-Attention.

\noindent\textbf{Effect of Patch-Level Weighting Module.} 
After introducing the PLW module, all three reference-based metrics show consistent improvement, indicating that this module effectively compensates for the deficiency of global attention in local modeling. By adaptively modulating the fusion weights based on the local context, the PLW module suppresses artifacts in changed regions while enhancing textures in unchanged areas. \cref{fig:ablation}(c) further corroborates this observation, as the texture details of buildings are noticeably sharper and more refined.

\noindent\textbf{Effect of Autoguidance.} 
When autoguidance is applied to the baseline model, all metrics improve, with particularly significant gains in perceptual and no-reference metrics. This indicates that the corrected velocity field steers the sampling trajectory toward data distributions with richer high-frequency details, thereby producing more realistic reconstructions.

\noindent\textbf{Complementarity of Components.} 
When both modules are introduced simultaneously, all metrics reach their optimal values, showing larger gains than introducing either module alone. This suggests that the PLW module and autoguidance are highly complementary. Specifically, the PLW module facilitates the recovery of local structures, while autoguidance enhances global textural realism; together, they improve the overall reconstruction quality. As shown in \cref{fig:ablation}(d), the reconstructed result maintains high fidelity while exhibiting stronger perceptual realism.

\begin{table}[tb]
    \caption{Ablation study on the SECOND dataset. The results are reported for $\times 8$ (upper four rows) and $\times 16$ (lower four rows) scaling factors. \checkmark indicates the component is included. The best result in each group is highlighted in \textbf{bold}.
    }
    \label{tab:ablation}
    \centering
    \begin{tabular}{@{}ccc ccccc@{}}
      \toprule
      Siamese & PLW & Auto- 
        & LPIPS$\downarrow$ &  FID$\downarrow$  & DISTS$\downarrow$ 
        & CLIPIQA$\uparrow$ & MUSIQ$\uparrow$ \\
      Attention & Module & guidance & & & & & \\
      \midrule
      \checkmark &            &            & 0.2362 & 19.16 & 0.1251 & 0.4403 & 44.95 \\
      \checkmark & \checkmark &            & 0.2291 & 18.93 & 0.1234 & 0.4399 & 45.07 \\
      \checkmark &            & \checkmark & 0.2278 & 18.56 & 0.1159 & 0.4680 & 48.29 \\
      \checkmark & \checkmark & \checkmark & \textbf{0.2174} & \textbf{15.87} & \textbf{0.1090} & \textbf{0.4698} & \textbf{49.04} \\
      \midrule
      \checkmark &            &            & 0.3159 & 31.29 & 0.1571 & 0.4514 & 43.78 \\
      \checkmark & \checkmark &            & 0.3085 & 30.32 & 0.1557 & 0.4604 & 43.88 \\
      \checkmark &            & \checkmark & 0.3060 & 27.96 & 0.1449 & 0.4649 & 46.60 \\
      \checkmark & \checkmark & \checkmark & \textbf{0.2905} & \textbf{21.58} & \textbf{0.1334} & \textbf{0.4875} & \textbf{48.65} \\
      \bottomrule
    \end{tabular}
\end{table}

\begin{table}[tb]
  \caption{Comparison of different injection strategies on the SECOND $\times 8$ dataset. The best result is highlighted in \textbf{bold}.
  }
  \label{tab:injection}
  \centering
  \begin{tabular}{@{}c ccccc@{}}
    \toprule
    Method & PSNR~$\uparrow$ & SSIM~$\uparrow$ & LPIPS~$\downarrow$ & FID~$\downarrow$ & DISTS~$\downarrow$ \\
    \midrule
    Siamese Attention      & 24.50 & 0.5921 & 0.2362 & 19.16 & 0.1251 \\
    + Variant A       & 24.63 & 0.5981 & 0.2327 & 23.07 & 0.1303 \\
    + Variant B       & 24.63 & 0.5966 & 0.2308 & 19.81 & 0.1254 \\
    + Variant C (PLW) & \textbf{24.70} & \textbf{0.6005} & \textbf{0.2291} & \textbf{18.93} & \textbf{0.1234} \\
    \bottomrule
  \end{tabular}
\end{table}

\subsection{Discussion on LR and Ref Injection Strategy}
\label{subsec:discussion}

It is worth noting that not all injection strategies lead to performance improvements. We initially designed two simpler injection approaches. Variant A keeps LR and Ref isolated from each other, avoiding their explicit interaction. The LR tokens $\mathbf{H}^l$ and Ref tokens $\mathbf{H}^r$ are injected into the noisy tokens $\mathbf{H}^z$ through independent zero-initialized linear layers:
\begin{equation}
    \tilde{\mathbf{H}}^z = \mathbf{H}^z + \mathrm{ZeroLinear}(\mathbf{H}^l) + \mathrm{ZeroLinear}(\mathbf{H}^r).
    \label{eq:variant_a}
\end{equation}
Variant B directly sums LR and Ref before injecting them into the noisy tokens through a single zero-initialized linear layer:
\begin{equation}
    \tilde{\mathbf{H}}^z = \mathbf{H}^z + \mathrm{ZeroLinear}(\mathbf{H}^l + \mathbf{H}^r).
    \label{eq:variant_b}
\end{equation}
Our proposed Patch-Level Weighting (PLW) injection strategy is denoted as Variant C, with the corresponding formulation given in \cref{eq:patch_fusion}. The comparison results on the SECOND $\times 8$ dataset are presented in \cref{tab:injection}. Variant A improves PSNR and SSIM but leads to a higher FID, suggesting that independently injecting LR and Ref information does not necessarily improve perceptual quality. Variant B reduces FID compared with Variant A, but its direct summation still provides limited gains over the baseline. In contrast, Variant C (PLW) achieves the best overall performance, indicating the benefit of adaptive per-patch fusion.

However, the design of the PLW module implicitly assumes that LR and Ref are roughly spatially aligned. This assumption generally holds in remote sensing scenarios, where both images originate from the same geographic location. For spatially misaligned reference images, \eg, captured from different viewpoints or covering different regions, the per-patch weight allocation would lose its spatial correspondence, potentially leading to degraded performance. How to adaptively retrieve and exploit useful texture information from spatially misaligned reference images remains an important direction for our future exploration.

\section{Conclusion}
In this work, we propose the Decoupled Siamese Diffusion Transformer (DS-DiT), a framework designed to balance LR structural fidelity and Ref texture utilization in remote sensing RefSR. Through the decoupled siamese attention mechanism, DS-DiT separates the interaction between LR and Ref conditions, preserving structural fidelity while incorporating high-frequency details. To further address the limitations of global attention, we design a Patch-Level Weighting module for adaptive local feature fusion. Additionally, we introduce an autoguidance strategy that improves reconstruction quality during inference by exploiting the model's internal structural priors without requiring additional training. This work demonstrates the significant potential of DS-DiT for high-fidelity remote sensing reconstruction, providing a robust and flexible approach for practical applications where precise texture recovery must be balanced with strict structural authenticity.

\par\vfill\par

\section*{Acknowledgements}
This research was supported in part by the Guangdong S\&T Program (Grant No. 2024B0101040005), the National Natural Science Foundation of China (Grant No. T2125006), and the Natural Science Foundation of Guangdong Province, China (Grant No. 2026A1515010676).
\bibliographystyle{splncs04}
\bibliography{main}

\appendix
\renewcommand{\theHsection}{appendix.\arabic{section}}
\section*{Appendix}
\section{Detailed Illustration of M$^3$-DiT}
\label{sec:appendix_m3dit}

\begin{figure}[h]
  \centering
  \includegraphics[width=\textwidth]{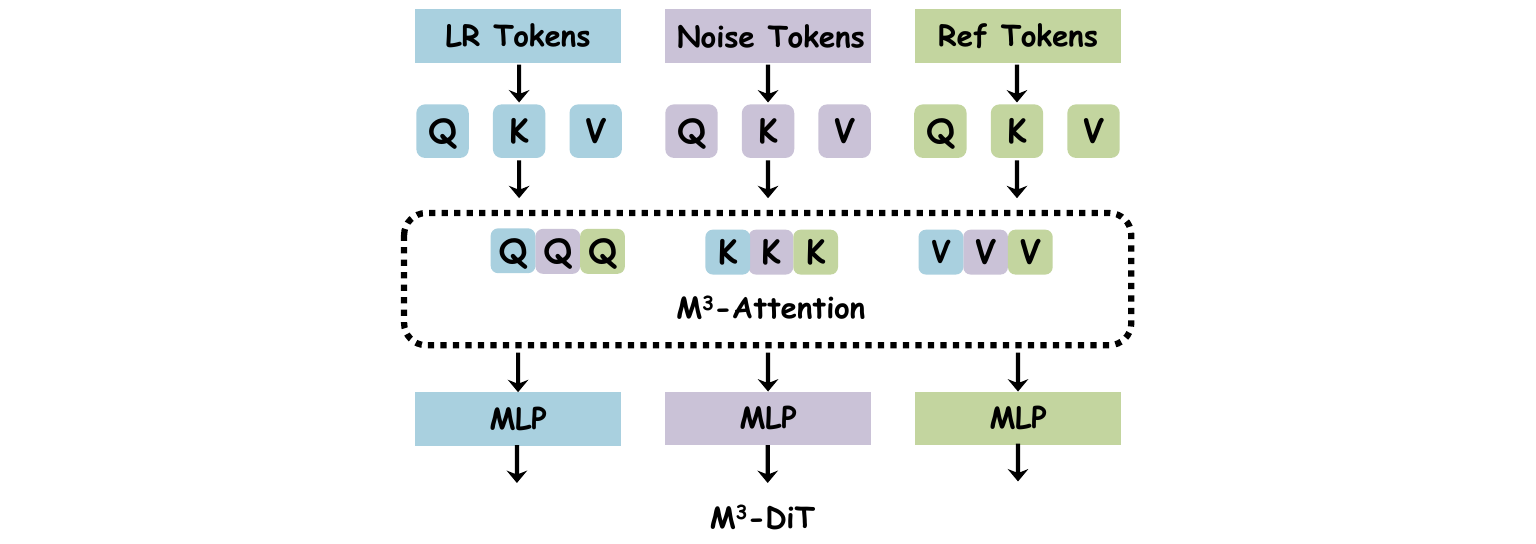}
  \caption{The architecture of M$^3$-DiT.}
  \label{fig:M3DiT}
\end{figure}

As mentioned in \cref{subsec:siamese_attention}, we adapt M$^3$-DiT for RefSR and include it as a comparison method. Here we provide a more detailed description. As illustrated in \cref{fig:M3DiT}, the noisy image tokens $\mathbf{h}^z$ produce a set of query, key, and value matrices in each block: $\mathbf{Q}^z = \mathbf{h}^z \mathbf{W}_q^z$, $\mathbf{K}^z = \mathbf{h}^z \mathbf{W}_k^z$, $\mathbf{V}^z = \mathbf{h}^z \mathbf{W}_v^z$, where $\mathbf{W}_q^z$, $\mathbf{W}_k^z$, and $\mathbf{W}_v^z$ are the weight matrices of the query, key, and value projection layers for $\mathbf{h}^z$. Likewise, the LR tokens $\mathbf{h}^l$ and Ref tokens $\mathbf{h}^r$ produce their respective queries, keys, and values: $\mathbf{Q}^l, \mathbf{K}^l, \mathbf{V}^l$ and $\mathbf{Q}^r, \mathbf{K}^r, \mathbf{V}^r$. The interaction among all three information sources is then realized through M$^3$-Attention:
\begin{equation}
  \mathbf{h}_{\mathrm{new}}^z,\;
  \mathbf{h}_{\mathrm{new}}^l,\;
  \mathbf{h}_{\mathrm{new}}^r =
  \mathrm{Attention}\!\left(
    [\mathbf{Q}^z, \mathbf{Q}^l, \mathbf{Q}^r],\;
    [\mathbf{K}^z, \mathbf{K}^l, \mathbf{K}^r],\;
    [\mathbf{V}^z, \mathbf{V}^l, \mathbf{V}^r]
  \right),
\end{equation}
where $[\cdot, \cdot, \cdot]$ denotes concatenation along the sequence dimension, and $\mathbf{h}_{\mathrm{new}}^z$, $\mathbf{h}_{\mathrm{new}}^l$, and $\mathbf{h}_{\mathrm{new}}^r$ are the updated tokens after interaction. Unlike our siamese attention design, M$^3$-DiT allows unconstrained interactions among all branches within a single unified attention sequence.

\section{Autoguidance Analysis}
\label{sec:appendix_autoguidance}

\subsection{Effect of Guidance Coefficient $\omega$}
\label{subsec:effect_omega}

As described in \cref{subsec:autoguidance}, the noisy image tokens in the siamese attention are updated via $\mathbf{h}_{\mathrm{new}}^z = \mathbf{h}_l^z + \lambda \mathbf{h}_r^z$. Autoguidance corrects the velocity field by extrapolating between the normal conditional prediction $\mathbf{v}_t^\theta(\mathbf{x}_t \mid c_{\mathrm{lr}}, c_{\mathrm{ref}})$ ($\lambda=1$) and the weakened conditional prediction $\mathbf{v}_t^{\theta^-}(\mathbf{x}_t \mid c_{\mathrm{lr}}, c_{\mathrm{ref}}^-)$ ($\lambda=0$), with the guidance coefficient $\omega$ controlling the extrapolation strength.

\begin{figure}[tb]
  \centering
  \includegraphics[width=\textwidth]{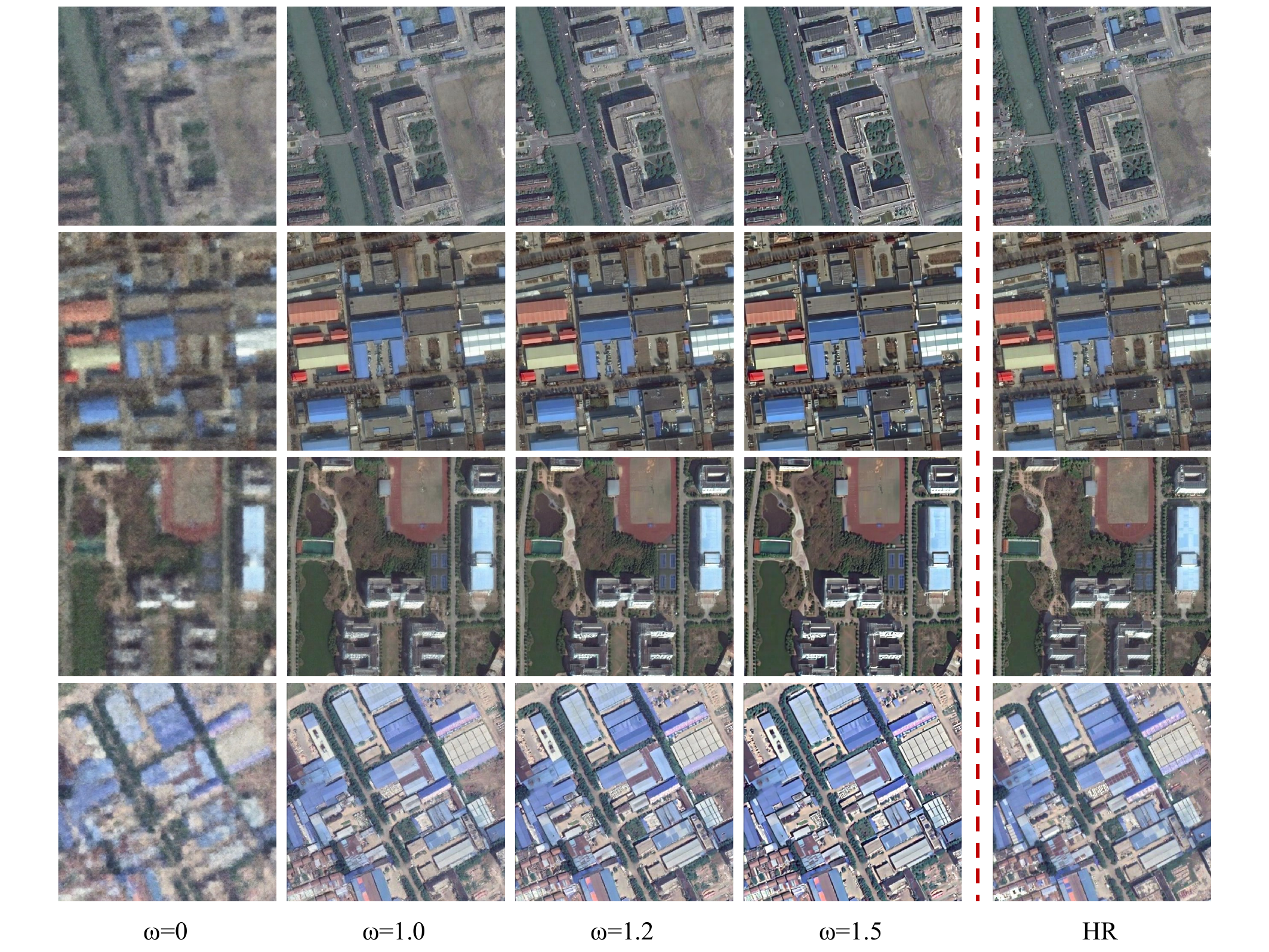}
  \caption{Impact of different guidance coefficients $\omega$ on the SECOND $\times 16$ dataset.}
  \label{fig:omega}
\end{figure}

To systematically analyze the effect of $\omega$ on reconstruction quality, we evaluate $\omega \in \{0, 1.0, 1.1, 1.2, 1.3, 1.4, 1.5\}$ on the SECOND dataset at both $\times 8$ and $\times 16$ scaling factors. The complete results are presented in \cref{tab:omega}. When $\omega = 0$, the prediction direction reduces to $\mathbf{v}_t^{\theta^-}$, where the direct contribution of the Ref branch to the noisy tokens in the attention stage is eliminated. As shown in \cref{fig:omega}, the resulting images are noticeably blurry and lack high-frequency details, with all metrics substantially worse than the baseline ($\omega = 1.0$). This confirms a significant directional discrepancy between the weakened and normal predictions, which is precisely the prerequisite for autoguidance to be effective.

As illustrated in \cref{fig:omega}, appropriately increasing $\omega$ within a certain range improves the visual quality of the reconstruction, yielding sharper land cover boundaries and finer details. However, excessively large $\omega$ leads to over-sharpening and color distortion, thereby degrading fidelity.

\begin{table}[tb]
  \caption{Effect of the guidance coefficient $\omega$ on reconstruction quality on the SECOND dataset at $\times 8$ and $\times 16$ scaling factors.
  }
  \label{tab:omega}
  \centering
  \setlength{\tabcolsep}{4pt}
  \begin{tabular}{@{}cc ccccccc@{}}
    \toprule
    Scale & $\omega$ & PSNR~$\uparrow$ & SSIM~$\uparrow$ & LPIPS~$\downarrow$ & FID~$\downarrow$ & DISTS~$\downarrow$ & CLIPIQA~$\uparrow$ & MUSIQ~$\uparrow$ \\
    \midrule
    \multirow{7}{*}{$\times 8$}
    & 0   & 23.11 & 0.5408 & 0.4465 & 75.14  & 0.2947 & 0.2691 & 23.47 \\
    & 1.0 & 24.70 & 0.6005 & 0.2291 & 18.93  & 0.1234 & 0.4399 & 45.07 \\
    & 1.1 & 24.48 & 0.5982 & 0.2193 & 16.62  & 0.1132 & 0.4595 & 47.27 \\
    & 1.2 & 24.17 & 0.5939 & 0.2174 & 15.87  & 0.1090 & 0.4698 & 49.04 \\
    & 1.3 & 23.78 & 0.5880 & 0.2223 & 16.72  & 0.1109 & 0.4724 & 50.50 \\
    & 1.4 & 23.31 & 0.5806 & 0.2323 & 18.88  & 0.1178 & 0.4668 & 51.73 \\
    & 1.5 & 22.79 & 0.5719 & 0.2459 & 22.14  & 0.1285 & 0.4541 & 52.80 \\
    \midrule
    \multirow{7}{*}{$\times 16$}
    & 0   & 21.11 & 0.4249 & 0.6201 & 159.66 & 0.4152 & 0.3044 & 18.69 \\
    & 1.0 & 22.16 & 0.4999 & 0.3085 & 30.32  & 0.1557 & 0.4604 & 43.88 \\
    & 1.1 & 21.86 & 0.4971 & 0.2939 & 24.14  & 0.1393 & 0.4790 & 46.56 \\
    & 1.2 & 21.48 & 0.4919 & 0.2905 & 21.58  & 0.1334 & 0.4875 & 48.65 \\
    & 1.3 & 21.01 & 0.4846 & 0.2960 & 22.04  & 0.1372 & 0.4848 & 50.28 \\
    & 1.4 & 20.49 & 0.4754 & 0.3078 & 25.01  & 0.1480 & 0.4713 & 51.57 \\
    & 1.5 & 19.92 & 0.4648 & 0.3235 & 29.97  & 0.1632 & 0.4473 & 52.61 \\
    \bottomrule
  \end{tabular}
\end{table}

\subsection{Comparison with Other Training-Free Guidance Methods}
\label{sec:supp_guidance_variants}

\begin{table}[tb]
  \caption{Comparison with other training-free guidance methods on SECOND $\times 8$.}
  \label{tab:guidance_variants}
  \centering
  \setlength{\tabcolsep}{4pt}
  \begin{tabular}{@{}c ccccc@{}}
    \toprule
    Method
    & LPIPS~$\downarrow$
    & FID~$\downarrow$
    & DISTS~$\downarrow$
    & CLIPIQA~$\uparrow$
    & MUSIQ~$\uparrow$ \\
    \midrule
    Baseline (w/o guidance)
    & 0.2291 & 18.93 & 0.1234 & 0.4399 & 45.07 \\

    + PAG
    & 0.2260 & \textcolor{second}{15.89} & \textcolor{second}{0.1106} & \textcolor{second}{0.4617} & \textcolor{best}{50.05} \\

    + SEG
    & \textcolor{second}{0.2192} & 15.97 & 0.1123 & 0.4597 & 46.58 \\

    + ICG
    & 0.2205 & 15.97 & 0.1135 & 0.4570 & 46.51 \\

    + AG (Ours)
    & \textcolor{best}{0.2174}
    & \textcolor{best}{15.87}
    & \textcolor{best}{0.1090}
    & \textcolor{best}{0.4698}
    & \textcolor{second}{49.04} \\
    \bottomrule
  \end{tabular}
\end{table}

We further compare the adopted autoguidance (AG) with three representative training-free guidance variants, including PAG~\cite{ahn2024self}, SEG~\cite{hong2024smoothed}, and ICG~\cite{sadat2025no}. For a fair comparison, the guidance scale of each variant is individually tuned to a competitive value. As shown in \cref{tab:guidance_variants}, our AG achieves a slight overall advantage over the compared guidance variants.

Training-free guidance methods usually construct a weak or perturbed prediction at inference time, which is not explicitly seen during training. PAG~\cite{ahn2024self} perturbs selected attention maps by replacing them with identity matrices, SEG~\cite{hong2024smoothed} smooths attention weights via Gaussian blurring, and ICG~\cite{sadat2025no} adopts independent out-of-distribution conditions. Importantly, these weak predictions are not used as standalone reliable outputs, but only to estimate a differential guidance direction. Our AG is in a similar spirit, but is tailored to the RefSR setting. The weak prediction is obtained by suppressing only the reference interaction branch, so it remains anchored to the LR structural prior while reducing the influence of reference information. This allows AG to provide an effective guidance direction without additional training.

\section{More Qualitative Comparisons}
\label{sec:supp_visual}

\begin{figure}[tb]
  \centering
  \includegraphics[width=\textwidth]{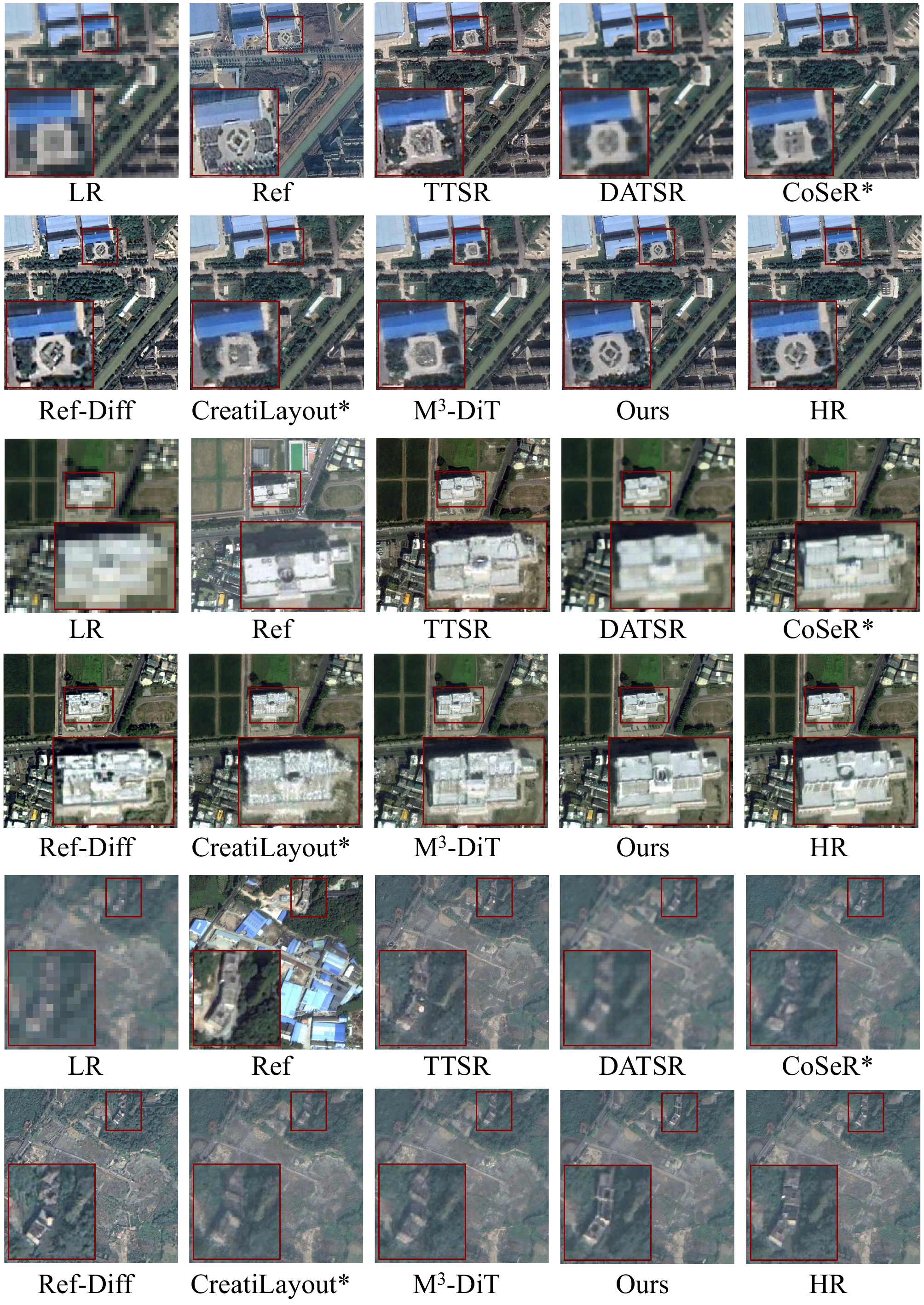}
  \caption{Qualitative comparison on the SECOND $\times 8$ dataset. 
  }
  \label{fig:qualitativeS1}
\end{figure}

\begin{figure}[tb]
  \centering
  \includegraphics[width=\textwidth]{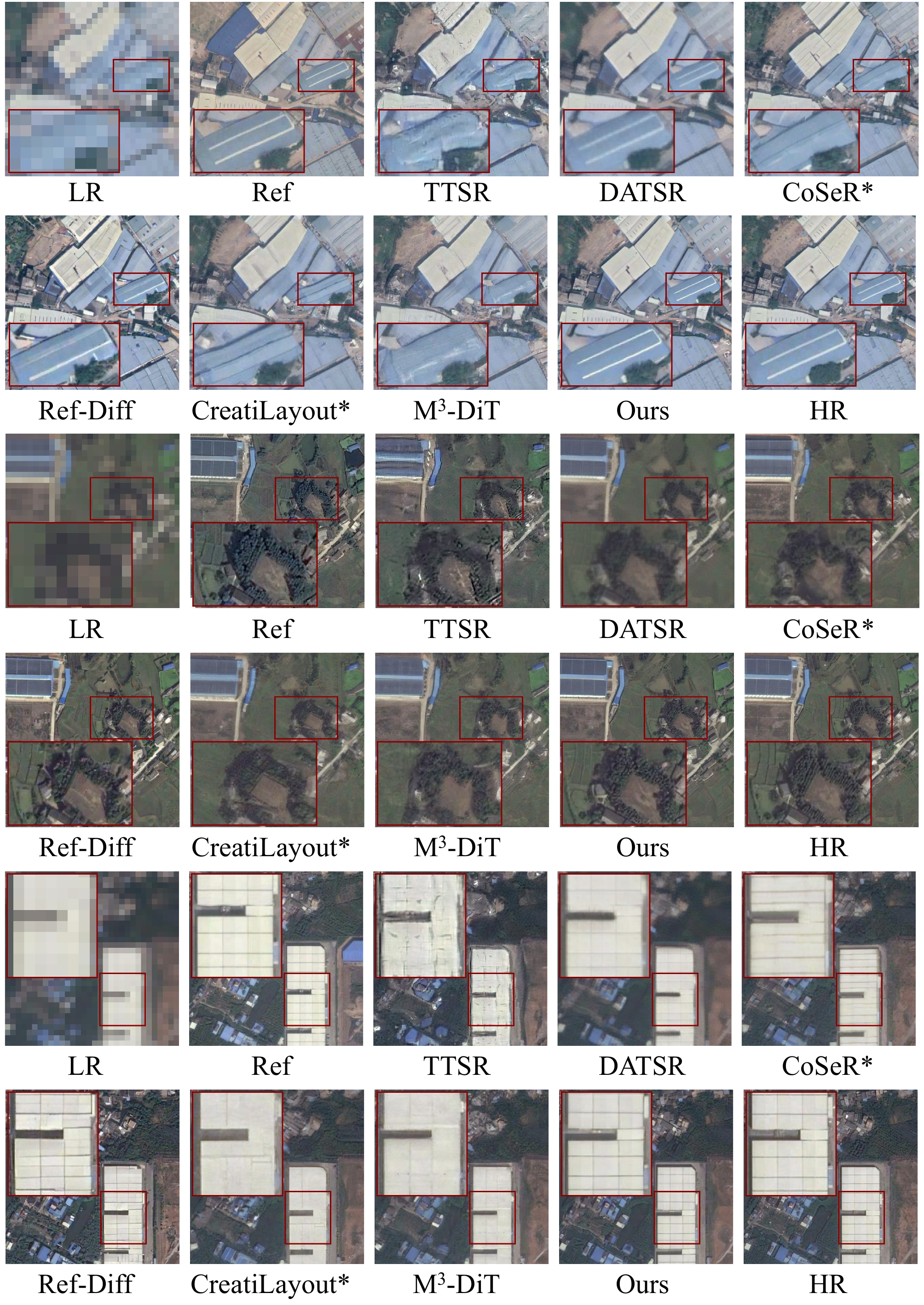}
  \caption{Qualitative comparison on the SECOND $\times 16$ dataset. 
  }
  \label{fig:qualitativeS2}
\end{figure}

\begin{figure}[tb]
  \centering
  \includegraphics[width=\textwidth]{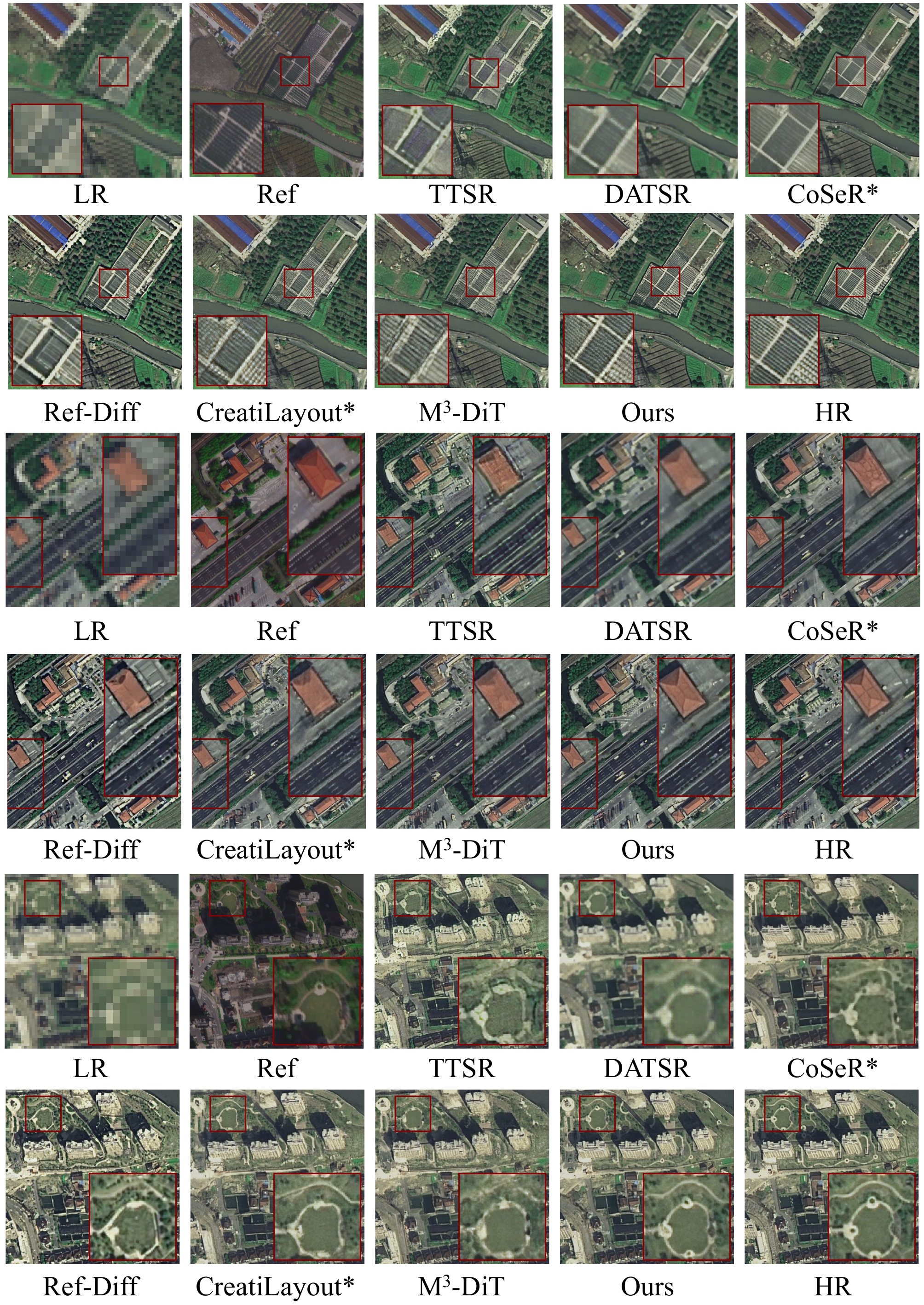}
  \caption{Qualitative comparison on the FUSU $\times 8$ dataset. 
  }
  \label{fig:qualitativeS3}
\end{figure}

\begin{figure}[tb]
  \centering
  \includegraphics[width=\textwidth]{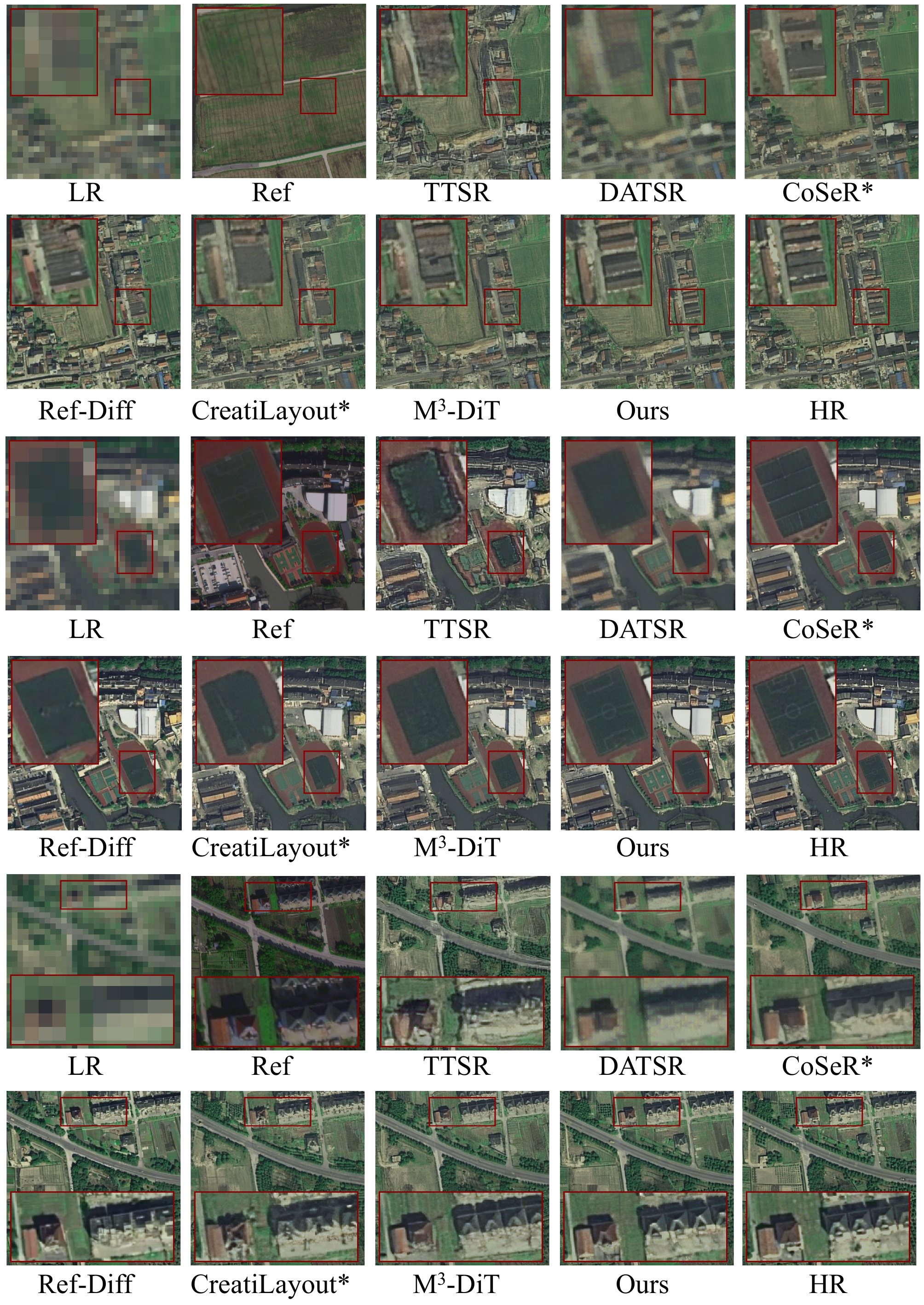}
  \caption{Qualitative comparison on the FUSU $\times 16$ dataset. 
  }
  \label{fig:qualitativeS4}
\end{figure}

We provide more visual comparisons of different RefSR methods on the SECOND~\cite{yang2020semantic} and FUSU~\cite{yuan2024fusu} datasets to further demonstrate the superiority of our approach. Specifically, \cref{fig:qualitativeS1,fig:qualitativeS2} display the super-resolution results on SECOND $\times 8$ and $\times 16$, respectively, while \cref{fig:qualitativeS3,fig:qualitativeS4} show the results on FUSU $\times 8$ and $\times 16$. As illustrated in these figures, our method consistently achieves superior visual quality across all settings.

\section{Computational Cost}
\label{sec:supp_computational_cost}

\begin{table}[h]
  \caption{Computational cost for generating a $512 \times 512$ image. Runtime is measured on a single NVIDIA A100 GPU.}
  \label{tab:efficiency}
  \centering
  \setlength{\tabcolsep}{4pt}
  \begin{tabular}{@{}c|cccc|cc@{}}
    \toprule
     & Ref-Diff & CoSeR$^{*}$ & Creati$^{*}$ & M$^3$-DiT & Ours & Ours (w/ AG) \\
    \midrule
    Params (M) 
    & 116.74 & 2655.50 & 3629.07 & 3005.81 
    & \multicolumn{2}{c}{3188.56} \\

    Steps 
    & 40 & 200 & 40 & 40 
    & \multicolumn{2}{c}{40} \\

    Time (s)     
    & 32.93 & 30.19 & 2.44 & 2.22 
    & 2.48 & 4.73 \\

    FLOPs (T)    
    & 940.63 & 249.93 & 93.46 & 81.85 
    & 90.88 & 181.75 \\
    \bottomrule
  \end{tabular}
\end{table}

We report the computational cost of diffusion-based RefSR methods when generating a $512 \times 512$ image. The runtime is measured on a single NVIDIA A100 GPU. As shown in \cref{tab:efficiency}, when AG is enabled, the inference cost increases because an additional weak prediction is computed for guidance, while the model parameters remain unchanged.

\section{Limitations} 
The coefficient $\omega$ in autoguidance remains a manually tuned hyperparameter, lacking an adaptive selection mechanism. 
Moreover, degradations in real-world scenarios are often complex and diverse~\cite{wang2021realesrgantrainingrealworldblind,lin2024diffbir,duan2025dit4sr}. Some Real-World Image Super-Resolution (Real-ISR) methods~\cite{yu2024scalingexcellencepracticingmodel,chen2024faithdiffunleashingdiffusionpriors,wu2024seesr} address this issue by training image encoders to extract useful features from degraded inputs. Integrating such strategies with our framework to handle real-world remote sensing scenarios is a promising direction for future work.

\end{document}